\documentclass[10pt]{arx}

\usepackage{tcolorbox}
\usepackage{xcolor}

\newtcolorbox{promptbox}{
  colback=gray!10,
  boxrule=0pt,
  arc=0pt,
  outer arc=0pt,
  left=5pt,
  right=5pt,
  top=5pt,
  bottom=5pt
}

\newenvironment{prompttext}
  {\ttfamily\small}
  {}

\title{Taxonomic Reasoning for Rare Arthropods:\\ 
Combining Dense Image Captioning and RAG for Interpretable Classification}

\author[1,2]{\small\bfseries Nathaniel Lesperance}
\author[3]{\small\bfseries Sujeevan Ratnasingham}
\author[1,2,3,*]{\small\bfseries Graham W.~Taylor}
\affil[1]{\small University of Guelph}
\affil[2]{\small Vector Institute for AI}
\affil[3]{\small Centre for Biodiversity Genomics}

\date{} 

\begin{document}
\maketitle
\renewcommand{\thefootnote}{\fnsymbol{footnote}} 
\footnotetext[1]{Author for correspondence: \texttt{gwtaylor@uoguelph.ca}}

\begin{abstract}
In the context of pressing climate change challenges and the significant biodiversity loss among arthropods, automated taxonomic classification from organismal images is a subject of intense research. However, traditional AI pipelines based on deep neural visual architectures such as CNNs or ViTs face limitations such as degraded performance on the long-tail of classes and the inability to reason about their predictions. We integrate image captioning and retrieval-augmented generation (RAG) with large language models (LLMs) to enhance biodiversity monitoring, showing particular promise for characterizing rare and unknown arthropod species. While a na\"ive Vision-Language Model (VLM) excels in classifying images of common species, the RAG model enables classification of rarer taxa by matching explicit textual descriptions of taxonomic features to contextual biodiversity text data from external sources. The RAG model shows promise in reducing overconfidence and enhancing accuracy relative to nai\"ve LLMs, suggesting its viability in capturing the nuances of taxonomic hierarchy, particularly at the challenging family and genus levels. Our findings highlight the potential for modern vision-language AI pipelines to support biodiversity conservation initiatives, emphasizing the role of comprehensive data curation and collaboration with citizen science platforms to improve species identification, unknown species characterization and ultimately inform conservation strategies.
\end{abstract}

\section{Introduction}
\label{intro}

The ongoing global climate change polycrisis contributes significantly to the loss of biodiversity worldwide, particularly in Arthropoda \citep{seibold_arthropod_2019}. Arthropods are found in every habitat on Earth, playing a myriad of ecological roles \citep{yang_insects_2014}. They contribute significantly to the health of a given ecosystem \citep{culliney_role_2013} and they are the most diverse phylum, making up approximately 84\% of known animal species \citep{zhang_animal_2013}. Despite their diversity, many more species are unknown than known to science \citep{hartop_resolving_2024, odegaard_how_2000}, making the monitoring of species diversity, distributions, and phenology challenging. Arthropods' overrepresentation among invasive species further threatens biodiversity conservation. The precise taxonomic classification of known and unknown species of arthropods is fundamental for biodiversity conservation and to limiting the negative impacts of climate change for all life on Earth.

\begin{figure}[ht]
  \centering
  \includegraphics[width=\textwidth]{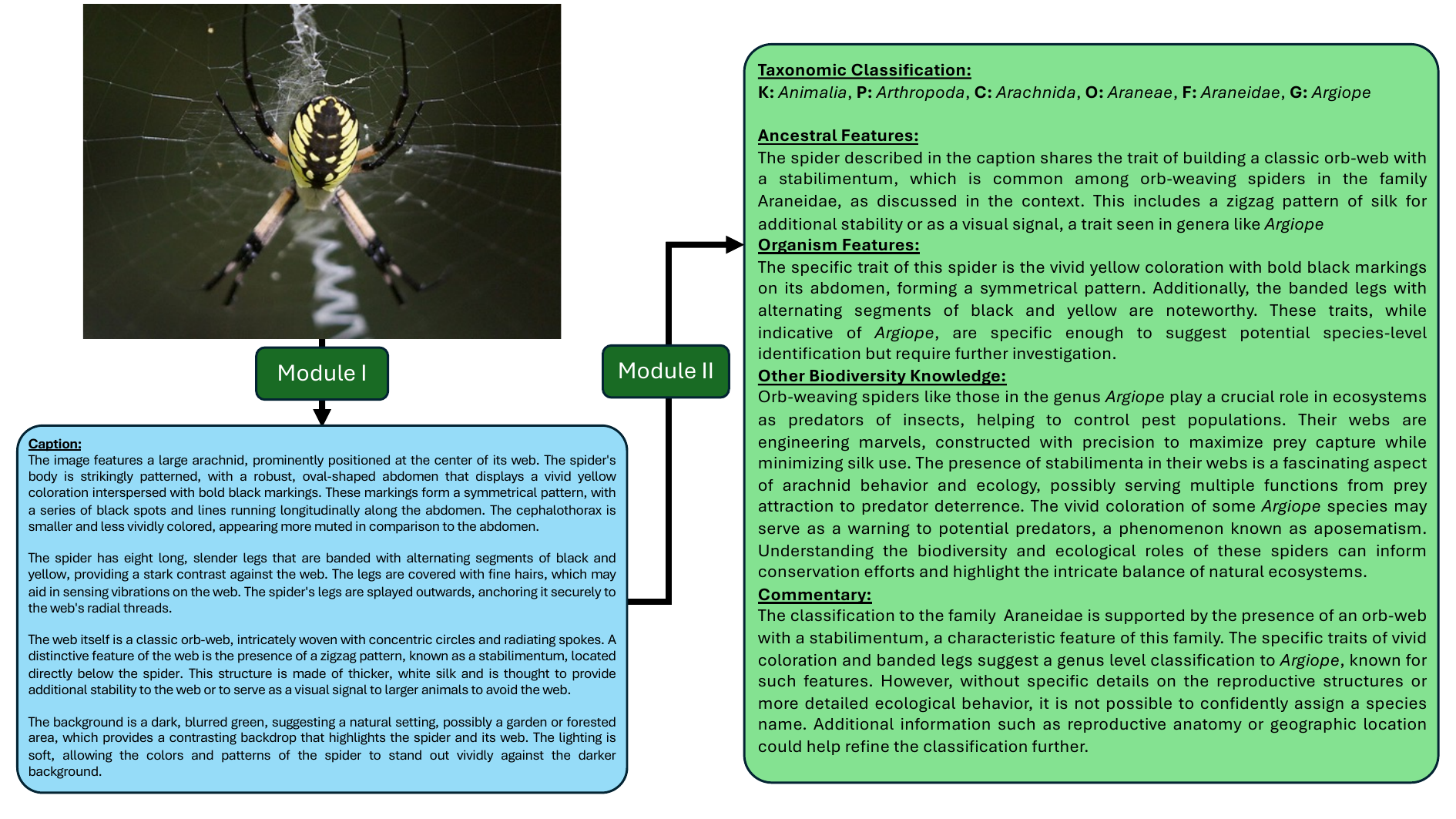}
  \caption{Example output from Module I (light blue) and Module II (light green) for an image of a yellow garden spider (Photo attribution: David Illig). In taxonomic classification: K=Kingdom, P=Phylum, C=Class, O=Order, F=Family, G=Genus. The RAG model generated accurate taxonomic labels to the genus level and refrained from providing a species level classification. }
  \label{fig:ex_output}
\end{figure}

Taxonomic classification using DNA-based methods is the primary driver to recent breakthroughs in biodiversity understanding, initiated by the characterization of DNA barcodes and the adoption of species-like clusters called Barcode Index Numbers (BINs) that dramatically expanded our biodiversity catalogue \citep{hebert_biological_2003, ratnasingham_BINs_2013}. Recent improvements in deep learning have provided new opportunities for automated DNA-based classification at scale such as the Nucleotide Transformer \citep{dalla-torre_nucleotide_2024} and HyenaDNA \citep{nguyen_hyenadna_2023}. Unimodal automated taxonomic systems optimized for and trained on DNA barcodes have emerged, including BarcodeBERT \citep{millan_arias_barcodebert_2023}, MycoAI \citep{romeijn_mycoai_2024} and BarcodeMamba \citep{gao_barcodemamba_2024}. \citet{jain_machine_2024} and \citet{bjerge_real-time_2022} trained unimodal automated taxonomic systems using images from camera traps. In multimodal approaches, CLIBD \citep{gong_clibd_2024} extends BioCLIP's \citep{stevens_bioclip_2023} contrastive learning of text and image embeddings to include DNA, addressing the difficulty of obtaining comprehensive expert taxonomic labels at fine scale. These two models and others \citep{badirli_classifying_2023} also mention the characterization of unknown species as a motivation for their architecture and learning strategies. These models have facilitated biodiversity monitoring at scale and reduce the manual labour of domain experts. Yet, to ecologists, they are effectively black boxes.\looseness=-1

Prior to DNA's widespread adoption, taxonomists relied on detailed textual descriptions and references to classify specimens and identify new species. In contrast, today's automated taxonomic systems typically process raw DNA sequences or images in isolation, without incorporating this rich contextual knowledge. Citizen science resources like iNaturalist \citep{inaturalist_database}, Wikipedia and Wikispecies are well maintained and continually updated, allowing them to play a key role in data collection and curation for biodiversity research. While DNA-based methods like Barcode Index Numbers (BINs) have expanded our biodiversity catalogue, newly discovered species often lack the dense trait descriptions that traditionally accompany taxonomic classification. The combination of untapped text-based biodiversity resources and the success of Large Language Models (LLMs) at text-based reasoning enables automated taxonomic classification using LLMs that leverage visual features. 

Retrieval Augmented Generation (RAG), which can provide additional context to LLMs at inference has been shown to help in overcoming knowledge limitations in training data, obfuscation of reasoning processes and hallucinations seen in LLM responses \citep{gupta_comprehensive_2024, lewis_retrieval-augmented_2021, bai_hallucination_2024}. 
In the biological domain, RAG has been deployed in exploration of protein-protein interactions \citep{li_grappi_2025} but there are no known applications of RAG in taxonomic classification. Additionally, multimodal LLMs show promise in augmenting human biological research \citep{zhang_multimodal_2024} and have been used to generate text captions of scientific figures \citep{kim_multi-llm_2025}. Images of living arthropods can provide additional context clues that assist in taxonomic classification like habitat, interactions and behaviour. These supplementary context clues are present in taxonomy literature for retrieval and mostly missing from specimen images common to large arthropod image datasets like TOL-10M \citep{stevens_bioclip_2023} and BIOSCAN-5M \citep{gharaee_bioscan-5m_2025}. The following exploration pairs multimodal LLMs as an image captioning tool with RAG systems to reason over curated biodiversity databases with the aim to generate taxonomic classifications and accelerated biodiversity knowledge for images of arthropods (\autoref{fig:ex_output}). 

Automated taxonomic classification of organismal images using captioning paired with RAG against a vector database built from rich text-based knowledge bases enables minimally invasive and scalable zero-shot classification at fine-grained taxonomic levels. Our approach brings a generalization benefit to moving beyond traditional fixed-class datasets, especially when many classes remain unknown, as is common in biodiversity research \citep{Collins2013-vm}. Combining computer vision, natural language processing, and biodiversity informatics can address taxonomic classification challenges, strengthen citizen science platforms, and support global biodiversity efforts like BIOSCAN \citep{BIOSCAN_2025} and Biodiversity Genomics Europe \citep{BGE_2024}.

\section{Methods}
\label{methods}

\begin{figure}[ht]
  \centering
  \includegraphics[width=\textwidth]{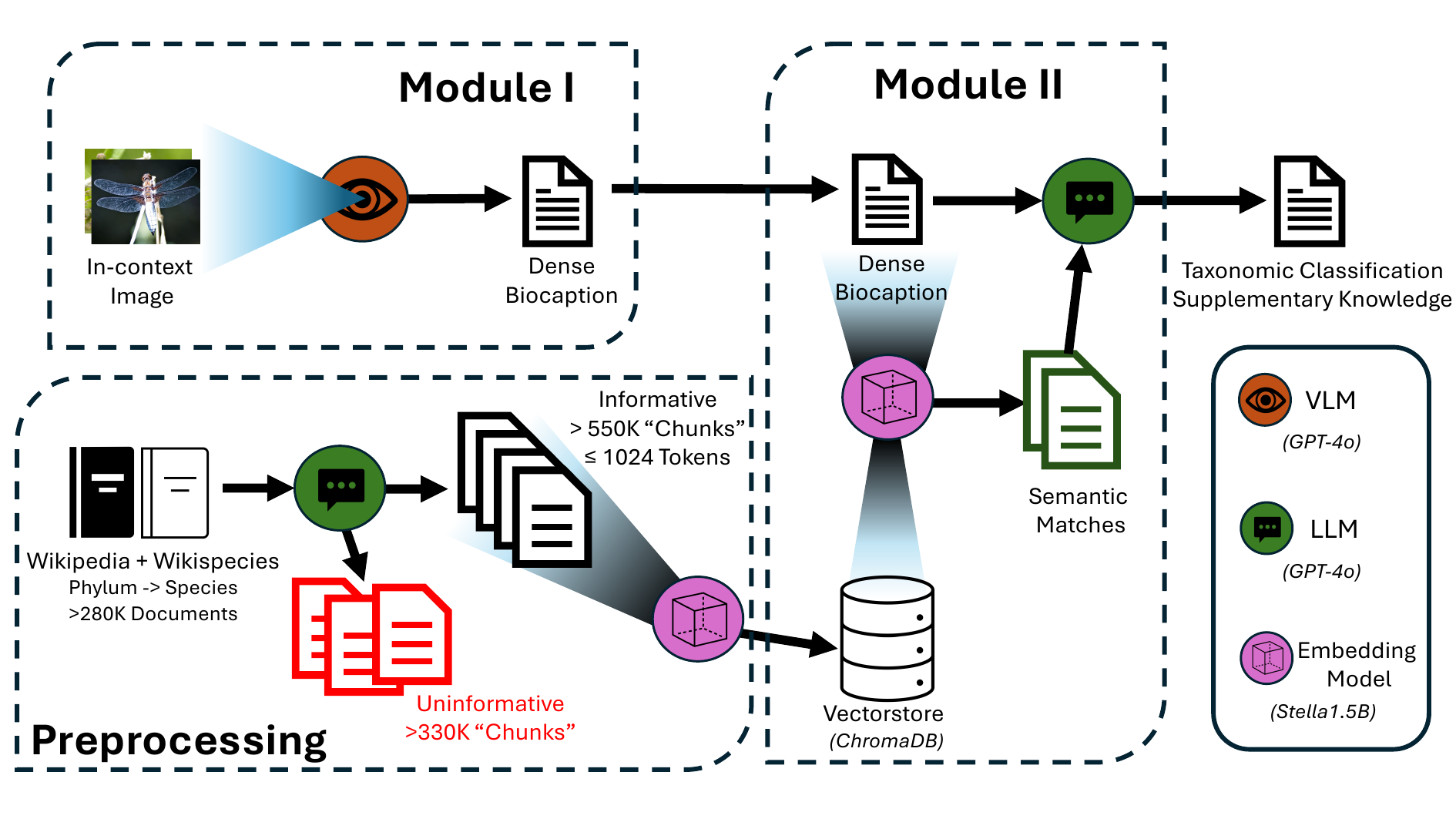}
  \caption{Preprocessing of Wikipedia and Wikispecies text data includes chunking, filtering and contextualization before generating a vector database of embeddings. In-context images of Arthropoda are inputs to the RAG model and are initially fed through a VLM (Module I) to generate dense image biocaptions which are the subject of RAG queries to the vector database in Module II. The RAG model generates taxonomic classifications, accelerated biodiversity knowledge and commentary on LLM confidence for each image.}
  \label{fig:pipeline_fig}
\end{figure}

\autoref{fig:pipeline_fig} summarizes our pipeline, detailing data source preprocessing, Module I - Image Captioning, and Module II - RAG System, used for generating taxonomic classifications and supplemental knowledge from organismal images. We now discuss each step in detail.

\subsection{Data Collection and Preprocessing}

Work with VLMs and LLMs is carried out using LangChain \citep{chase2022langchain}, instructor \citep{liu_instructor_2023}, and custom prompts (\autoref{appendix-contextualize prompt}, \autoref{appendix-captioning prompt}, \autoref{appendix-rag prompt}) that also contain thinking dots \citep{pfau_lets_2024}.\looseness=-1

\subsubsection{Image Datasets}

\noindent \textbf{Living Arthropods:} We curated a development dataset of 240 arthropod images photographed primarily in their natural habitats from Wikimedia and Ontario Nature Magazine. We used this dataset to validate our approach and optimize hyperparameters while managing computational costs. Every image has comprehensive taxonomic labels spanning 5 classes, 18 orders, 65 families, 120 genera, and 153 species within the Phylum Arthropoda with significant Class Insecta representation reflecting typical phylum distributions.

\noindent \textbf{Rare Species subset:} We selected all 951 Arthropoda images from the Rare Species dataset \citep{stevens_bioclip_2023}, including both in-situ habitat photographs and laboratory specimen images. These species are classified by the \href{https://www.iucnredlist.org}{IUCN Red List} \citep{iucn_2024} as Near Threatened, Vulnerable, Endangered, Critically, Endangered, or Extinct in the Wild. Each image has complete taxonomic labels across 4 classes, 9 orders, 17 families, 26 genera, and 32 species within the Phylum Arthropoda, again with Class Insecta being most represented. We provide detailed taxonomic counts for both datasets in \autoref{appendix-dataset tables}.

\subsubsection{Text-based Knowledge Sources}

We tokenized all text from Wikipedia and Wikispecies articles within the Kingdom Animalia, (\num{280120} documents) using Byte-Pair Encoding and recursively split it into chunks using whitespace as delimiters, generating \num{879611} total starting chunks. We then performed semantic filtering of uninformative chunks, reducing the total to \num{550711} chunks. For each remaining chunk, we used custom LLM prompts (\autoref{appendix-contextualize prompt}) to generate contextualizing text~\citep{anthropic2024contextual}, processing each chunk along with its full corresponding source document where all chunks post-processing were <= 1024 tokens.

\subsubsection{RAG Vectorstore}

We generated embeddings with ChromaDB \citep{chroma_chromadb_2022} using the \texttt{\seqsplit{stella\_en\_1.5B\_v5}} model \citep{zhang2025jasperstelladistillationsota}. We normalized all embedding vectors to length unity and tagged each chunk with metadata containing taxonomic rank, data source (Wikipedia or Wikispecies) and taxonomic name.

\subsection{Module I - Image Captioning}

Traditional captioners are trained to generate short descriptions for generic scenes and lack the detail required to produce detailed biologically-informed captions (biocaptions). Image-caption datasets are also generic and there are no known datasets geared towards this producing biocaptions. To pair well with the RAG task, biocaptions should exclusively describe visible features without inappropriate inference that could cloud taxonomic classification. Instead, each image in the dataset is passed to OpenAI's multimodal GPT-4o model \citep{openai_gpt-4o_2024} with custom prompting for a dense descriptive caption that could be informative for the taxonomic classification task (\autoref{appendix-captioning prompt}), including all visible features of the primary organism and its surroundings. Negative prompting helps prevent unnecessary inference about taxonomic classification, unseen behaviour or non-visible body parts. 

\subsection{Module II - RAG System}
\label{RAG_system}

\noindent \textbf{Query construction:} We used the dense descriptive captions from Module I to query the vector database. Our querying process begins by generating a normalized semantic embedding query vector using the same \texttt{\seqsplit{stella\_en\_1.5B\_v5}} embedding model. 

\noindent \textbf{Retrieval:} We built two retrieval pipelines to test the impact of advanced retrieval methods on model performance. In the `Simple RAG' model, we use dense search, maximizing cosine similarity to retrieve chunks that semantically match the query. We use default search parameters except for \textit{top k} (30 chunks per query). In the `Advanced RAG' model, we use multiquery retrieval with an instance of GPT-4o-mini to generate multiple artificial semantically similar queries. We use dense search selecting by Maximal Marginal Relevance (MMR) to obtain diverse chunks matching these queries. The set of unique chunks retrieved is passed to Cohere's reranking model \texttt{\seqsplit{rerank-english-v3.0}} \citep{cohere_reranker_2024} and the top 10 chunks are provided as context for response generation. 

\noindent \textbf{Response generation:} We passed the dense descriptive caption of the image, along with the text and metadata from the retrieved chunks, to another instance of GPT-4o with a custom prompt (\autoref{appendix-rag prompt}). This prompt instructed the model to generate taxonomic classifications at each rank, but only when confident in its classifications. We instructed the model to refrain from making classifications at or below any rank where the evidence in the caption and context was insufficient. In the same pass, we also prompted the model to generate biodiversity knowledge for the organism based on the caption, context and classification, along with commentary on its confidence and choices.

\subsubsection{Na\"ive VLMs and Na\"ive LLM}

To evaluate the contribution of each component in our pipeline, we implemented two baseline variants. Another instance of the same GPT-4o model used in Module I (known here as `Na\"ive VLM') generates responses from images alone without Module I's captioning or Module II's retrieval of additional context from the database. Another instance of the same GPT-4o model used in Module II (known here as `Na\"ive LLM') generates responses directly from dense captions alone without retrieval of additional context from the database. In evaluations on the Rare Species subset, we test the `Na\"ive VLM' and an instance of Google DeepMind's Gemini 2.0 Flash \citep{google_gemini_2024} (known here as `Na\"ive Gemini 2.0 Flash'). As a result, `Na\"ive VLM' is replaced with `Na\"ive GPT-4o' to disambiguate. 

\subsection{Evaluation}

\subsubsection{Classification}

To evaluate our hypothesis that RAG-based classification would perform better on rare taxa by leveraging external knowledge, we organized the Living Arthropod samples into rare taxa and common taxa based on the cumulative distribution of number of species observations ($n_\text{obs}$) in the iNaturalist database \citep{inaturalist_observations}. \textbf{Rare taxa} ($n = 109$) with $n_\text{obs}$ above the 95th percentile of the cumulative distribution ($n_\text{obs} < \num{6 699}$) and \textbf{common taxa} (n = 24) with $n_\text{obs}$ below the median of the cumulative distribution, ($n_\text{obs} > \num{60 562}$). We then evaluated classification macro accuracy of predicted labels at each taxonomic level for each of the Na\"ive VLM, Na\"ive LLM and Simple RAG Model. Since models were instructed to refrain from classification when uncertain, we also tracked the number of classification attempts made by each model --- this helps distinguish models that achieve high accuracy by being selective from those that maintain accuracy while attempting more classifications.

\subsubsection{RAG}

Retrieval augmented generation of accelerated biodiversity knowledge was evaluated using two custom measures: answer relevancy and faithfulness, provided by RAGAS \citep{es_ragas_2023} that use instances of GPT-4o-mini. We compared the Simple RAG model with the Advanced RAG model (both described in \autoref{RAG_system}). 

To assess \textbf{answer relevancy}, an LLM generates artificial questions based on the model's final output, then takes the mean of cosine similarities between embeddings of each artificial question and the model input query. To assess \textbf{faithfulness}, an LLM determines the total number of factual claims in the response and the number of context-supported claims. The ratio of context-supported claims to total claims is the faithfulness score. Both of these measures generate a score between 0 and 1 from each response. 

\section{Results}
\label{results}

Example output from Module I and the Module II (Simple RAG) is shown in \autoref{fig:ex_output}. 

\begin{figure}[ht]
  \centering
  \includegraphics[width=\textwidth]{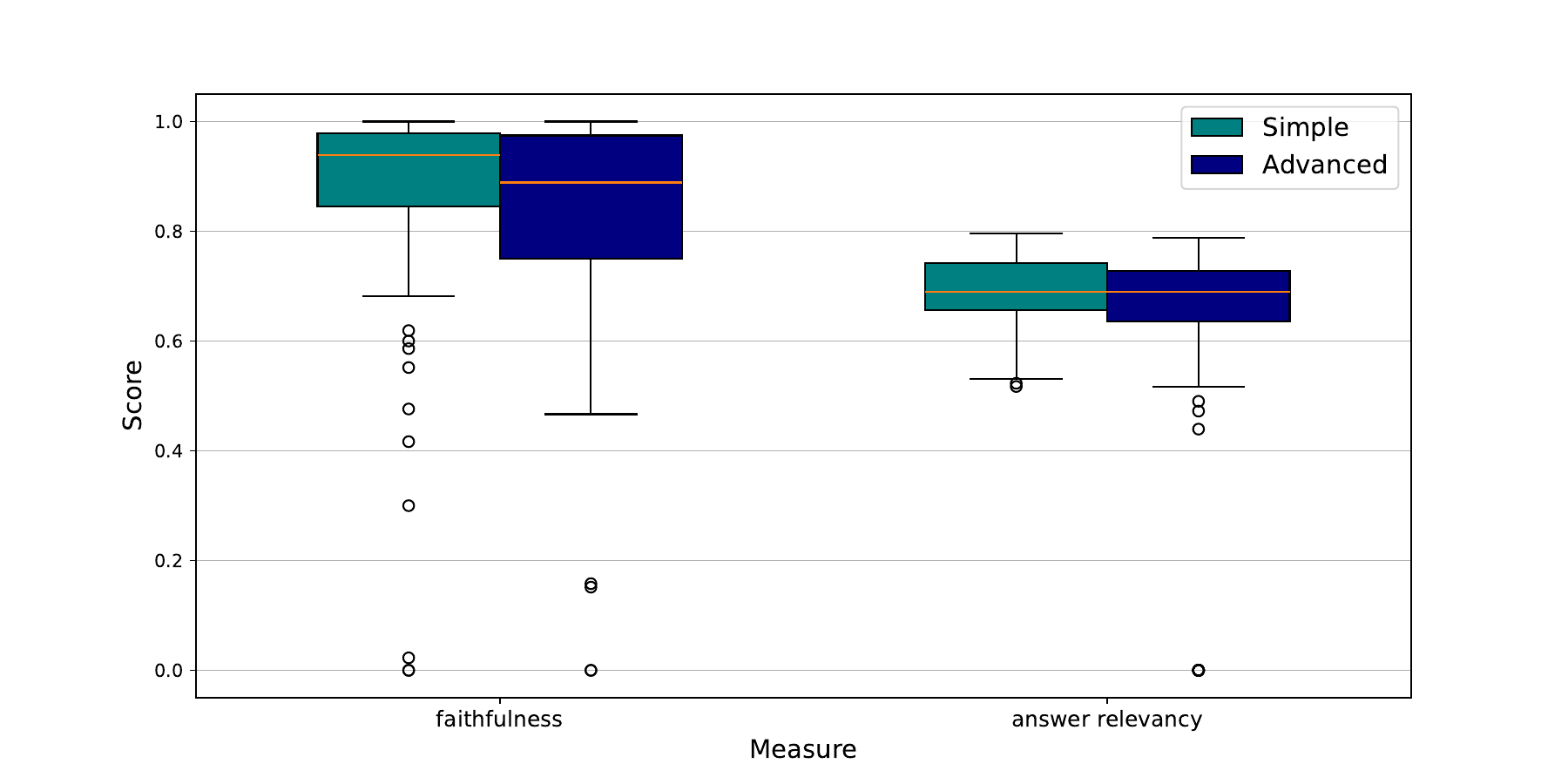}
  \caption{Faithfulness and answer relevancy score distributions for accelerated biodiversity knowledge for Simple and Advanced RAG models. Medians (red lines), Inter quartile ranges (colour-filled boxes), maxima, minima as well as outliers more than 1.5 IQR outside the 1st and 3rd quartiles are shown. The Advanced RAG model uses MMR search criterion, multi-query and reranking while the Simple RAG model uses cosine similarity search criterion, no multiquery and no reranking.}
  \label{fig:rag_eval_fig}
\end{figure}

\subsection{Validation: Image Captioning}
Module I was able to sufficiently describe visible features of arthropods including informed morphology, behaviour and contextualization. However, despite negative prompting, not all speculation or inferences about function not immediately evident in the image were eliminated during captioning. For example, in an example caption (\autoref{fig:ex_output}) the description ``...and is thought to provide additional stability to the web or to serve as a visual signal to larger animals to avoid the web'' is inappropriate speculation about the function that should be left to Module II during context comparison. 

\subsection{Validation: Retrieval Augmented Generation}

\autoref{fig:rag_eval_fig} shows distributions of the two key RAG metrics: answer relevancy and faithfulness, for the Simple RAG and Advanced RAG models. Both methods maintain high median faithfulness and answer relevancy. Multiple ablations of advanced RAG methods like reranking, alternative search criterion, and multi-query as well as variation in hyperparameters showed minimal impact on RAG performance (\autoref{fig:rag_eval_fig}) indicating a simpler RAG system was sufficient to maximize context retrieval of data available in the vector database. 

High median faithfulness and median answer relevancy (\autoref{fig:rag_eval_fig}) with minimal outliers indicates the generated biodiversity knowledge aligns with the assigned task and factually aligns with the retrieved context. This supports effective retrieval of context as well as effective text generation based on context.

\begin{table}[H]
  \centering
  \caption{Model classification attempts and classification macro accuracy (Acc.) for the Simple RAG model, Na\"ive LLM and Na\"ive VLM. Best performing models at each rank are in \textbf{bold} unless all models performed equally. Second best model is underlined.}
  \label{tab:main}
  
  \begin{subtable}{\textwidth}
    \centering
    \caption{All 240 arthropod images.}
    \label{tab:dev-primary}
\begin{tabular}{l|cS[table-format=1.4]|cS[table-format=1.4]|cS[table-format=1.4]}
\toprule
& \multicolumn{2}{c|}{Phylum} & \multicolumn{2}{c|}{Class} & \multicolumn{2}{c}{Order} \\
Model & Attempts (\%) & {Acc.} & Attempts (\%) & {Acc.} & Attempts (\%) & {Acc.} \\
\midrule
Simple RAG & 240 (100\%) & \num{1.00} & 240 (100\%) & \num{0.971} & 217 (90.4\%) & \uline{\num{0.977}} \\
Naïve LLM & 240 (100\%) & \num{1.00} & 240 (100\%) & \num{0.971} & 225 (93.8\%) & \num{0.942} \\
Naïve VLM & 240 (100\%) & \num{1.00} & 240 (100\%) & \num{0.971} & 240 (100\%) & \textbf{\num{0.983}} \\
\toprule
& \multicolumn{2}{c|}{Family} & \multicolumn{2}{c|}{Genus} & \multicolumn{2}{c}{Species} \\
Model & Attempts (\%) & {Acc.} & Attempts (\%) & {Acc.} & Attempts (\%) & {Acc.} \\
\midrule
Simple RAG & 35 (14.6\%) & \uline{\num{0.657}} & 3 (1.3\%) & \textbf{\num{1.00}} & 0 (0\%) & \text{--} \\
Naïve LLM & 61 (25.4\%) & \num{0.541} & 11 (4.6\%) & \num{0.364} & 0 (0\%) & \text{--} \\
Naïve VLM & 240 (100\%) & \textbf{\num{0.746}} & 240 (100\%) & \uline{\num{0.442}} & 230 (95.8\%) & \num{0.248} \\
\bottomrule
\end{tabular}
  \end{subtable}
  
  \vspace{1em} 
  
  \begin{subtable}{0.46\textwidth}
    \centering
    \caption{Common taxa ($n = 24$, $n_\text{obs} > 60 562$).}
    \label{tab:dev-secondary-common}
\setlength{\tabcolsep}{5pt}
\resizebox{\textwidth}{!}{%
\begin{tabular}{l|cS[table-format=1.4]|cS[table-format=1.4]|cS[table-format=1.4]}
\toprule
& \multicolumn{2}{c|}{Phylum} & \multicolumn{2}{c|}{Class} & \multicolumn{2}{c}{Order} \\
Model & Attempts (\%) & {Acc.} & Attempts (\%) & {Acc.} & Attempts (\%) & {Acc.} \\
\midrule
Simple RAG & 24 (100\%) & \num{1.00} & 24 (100\%) & \num{1.00} & 24 (100\%) & \num{1.00} \\
Naïve LLM & 24 (100\%) & \num{1.00} & 24 (100\%) & \num{1.00} & 24 (100\%) & \num{1.00} \\
Naïve VLM & 24 (100\%) & \num{1.00} & 24 (100\%) & \num{1.00} & 24 (100\%) & \num{1.00} \\
\toprule
& \multicolumn{2}{c|}{Family} & \multicolumn{2}{c|}{Genus} & \multicolumn{2}{c}{Species} \\
Model & Attempts (\%) & {Acc.} & Attempts (\%) & {Acc.} & Attempts (\%) & {Acc.} \\
\midrule
Simple RAG & 5 (20.8\%) & \num{0.600} & 1 (4.2\%) & \textbf{\num{1.00}} & 0 (0\%) & \text{--} \\
Naïve LLM & 12 (50.0\%) & \uline{\num{0.750}} & 3 (12.5\%) & \num{0.667} & 0 (0\%) & \text{--} \\
Naïve VLM & 24 (100\%) & \textbf{0.958} & 24 (100\%) & \uline{\num{0.729}} & 24 (100\%) & \num{0.792} \\
\bottomrule
\end{tabular}
}
  \end{subtable}
  \hspace{1em} 
  \begin{subtable}{0.46\textwidth}
    \centering
    \caption{Rare taxa ($n = 109$, $n_\text{obs} > 6 699$).}
    \label{tab:dev-secondary-rare}
\setlength{\tabcolsep}{5pt}
\resizebox{\textwidth}{!}{%
\begin{tabular}{l|cS[table-format=1.4]|cS[table-format=1.4]|cS[table-format=1.4]}
\toprule
& \multicolumn{2}{c|}{Phylum} & \multicolumn{2}{c|}{Class} & \multicolumn{2}{c}{Order} \\
Model & Attempts (\%) & {Acc.} & Attempts (\%) & {Acc.} & Attempts (\%) & {Acc.} \\
\midrule
Simple RAG & 109 (100\%) & \num{1.00} & 109 (100\%) & \num{0.982} & 99 (90.8\%) & \uline{\num{0.970}} \\
Naïve LLM & 109 (100\%) & \num{1.00} & 109 (100\%) & \num{0.982} & 100 (91.7\%) & \num{0.930} \\
Naïve VLM & 109 (100\%) & \num{1.00} & 109 (100\%) & \num{0.982} & 109 (100\%) & \textbf{0.982} \\
\toprule
& \multicolumn{2}{c|}{Family} & \multicolumn{2}{c|}{Genus} & \multicolumn{2}{c}{Species} \\
Model & Attempts (\%) & {Acc.} & Attempts (\%) & {Acc.} & Attempts (\%) & {Acc.} \\
\midrule
Simple RAG & 18 (16.5\%) & \uline{\num{0.556}} & 2 (1.8\%) & \textbf{1.00} & 0 (0\%) & \text{--} \\
Naïve LLM & 24 (22.0\%) & \num{0.458} & 3 (2.8\%) & \num{0} & \num{0} (0\%) & \text{--} \\
Naïve VLM & 109 (100\%) & \textbf{0.651} & 109 (100\%) & \uline{\num{0.275}} & 109 (100\%) & \num{0.028} \\
\bottomrule
\end{tabular}
}
  \end{subtable}
\end{table}

\subsection{Validation: Taxonomic Classification}

\autoref{tab:dev-primary} shows classification accuracy as well as number of taxonomic classification attempts made by the Na\"ive VLM, Na\"ive LLM, and RAG models at each taxonomic rank in the Living Arthropod dataset. The RAG model and Na\"ive VLM perform similarly at the phylum, class, and order levels, both achieving higher accuracy and lower overconfidence than the Na\"ive LLM. At the family level, Na\"ive VLM outperforms the RAG model with both exceeding the Na\"ive LLM in accuracy. The RAG model obtains perfect performance on its three genus-level attempts but is likely too conservative. At the species level, only the Na\"ive VLM makes classification attempts and shows a drop in accuracy despite little change to its confidence. For rare taxa (\autoref{tab:dev-secondary-rare}), however, the RAG model maintained its performance better than both Na\"ive LLM and Na\"ive VLM, which showed significant accuracy drops compared to their performance on common species (\autoref{tab:dev-secondary-common}). 

\subsection{Evaluation on IUCN Rare Species Dataset}
Having validated our approach on the smaller development dataset, we conducted our primary evaluation on the more challenging IUCN Rare Species Dataset (Arthropod subset). \autoref{tab:rs-performance} shows classification attempts, accuracy and F1 scores for the Simple RAG, Advanced RAG, Na\"ive GPT-4o and Na\"ive Gemini 2.0 Flash models on all 951 rare species images. Both RAG models consistently outperform both Na\"ive VLMs at the family, genus and species level. Both RAG pipelines show higher F1 scores compared to accuracy at the family, genus and species level, indicating elevated performance on minority classes. The Advanced RAG model's more sophisticated retrieval techniques (multiquery, MMR, and reranking) did not yield significant improvements over the Simple RAG model, suggesting that simpler retrieval methods are sufficient for the available knowledge base.

\begin{table}[t]
\centering
\caption{Model classification attempts, classification macro accuracy (Acc.) and F1 scores for the Simple RAG model, Advanced RAG model, Na\"ive GPT-4o and Na\"ive Gemini 2.0 Flash on 951 Rare Species arthropod images. Best performing models at each rank are in \textbf{bold} unless all models performed equally. Second best model is underlined. We count accuracy and F1 scores within $\pm~0.001$ as ties.}
\label{tab:rs-performance}
\small
\setlength{\tabcolsep}{4pt} 
\resizebox{\textwidth}{!}{%
\begin{tabular}{l|cS[table-format=1.4]@{\quad}S[table-format=1.4]|cS[table-format=1.4]@{\quad}S[table-format=1.4]|cS[table-format=1.4]@{\quad}S[table-format=1.4]}
\toprule
& \multicolumn{3}{c|}{Phylum} & \multicolumn{3}{c|}{Class} & \multicolumn{3}{c}{Order} \\
Model & Attempts (\%) & {Acc.} & {F1} & Attempts (\%) & {Acc.} & {F1} & Attempts (\%) & {Acc.} & {F1} \\
\midrule
Simple RAG & 936 (98.4\%) & \textbf{\num{0.994}} & \textbf{\num{0.997}} & 928 (97.6\%) & \num{0.976} & \uline{\num{0.979}} & 860 (90.4\%) & \num{0.900} & \num{0.919} \\
Advanced RAG & 932 (98.0\%) & \textbf{\num{0.995}} & \textbf{\num{0.997}} & 926 (97.4\%) & \num{0.976} & \uline{\num{0.978}} & 835 (87.8\%) & \uline{0.923} & \uline{0.935} \\
Naïve GPT-4o & 922 (96.9\%) & \uline{\num{0.985}} & \uline{\num{0.992}} & 922 (96.9\%) & \num{0.976} & \textbf{\num{0.983}} & 921 (96.9\%) & \num{0.921} & \num{0.933} \\
Naïve Gemini 2.0 Flash & 949 (99.8\%) & \uline{\num{0.985}} & \uline{\num{0.993}} & 949 (99.8\%) & \num{0.976} & \textbf{\num{0.984}} & 948 (99.7\%) & \textbf{\num{0.929}} & \textbf{\num{0.942}} \\
\midrule
& \multicolumn{3}{c|}{Family} & \multicolumn{3}{c|}{Genus} & \multicolumn{3}{c}{Species} \\
Model & Attempts (\%) & {Acc.} & {F1} & Attempts (\%) & {Acc.} & {F1} & Attempts (\%) & {Acc.} & {F1} \\
\midrule
Simple RAG & 419 (44.1\%) & \textbf{\num{0.928}} & \textbf{\num{0.934}} & 89 (9.4\%) & \uline{\num{0.843}} & \uline{\num{0.904}} & 43 (4.5\%) & \textbf{\num{0.512}} & \textbf{\num{0.547}} \\
Advanced RAG & 420 (44.1\%) & \uline{\num{0.924}} & \uline{\num{0.927}} & 82 (8.6\%) & \textbf{\num{0.854}} & \textbf{\num{0.919}} & 47 (4.9\%) & \uline{\num{0.468}} & \uline{\num{0.544}} \\
Naïve GPT-4o & 849 (89.3\%) & \num{0.873} & \num{0.909} & 459 (48.3\%) & \num{0.534} & \num{0.598} & 159 (16.7\%) & \num{0.157} & \num{0.223} \\
Naïve Gemini 2.0 Flash & 944 (99.3\%) & \num{0.798} & \num{0.836} & 763 (80.2\%) & \num{0.329} & \num{0.406} & 369 (38.8\%) & \num{0.195} & \num{0.265} \\
\bottomrule
\end{tabular}%
}
\end{table}

\section{Discussion}

In \autoref{tab:dev-primary} we see that the classification performance of the RAG model relative to the Na\"ive LLM shows reasoning over the additional retrieved context provides a small boost to LLM confidence and a modest gain in classification accuracy, particularly at the family and genus level where it is most needed. However, the improved classification performance of the Na\"ive VLM indicates many taxa present in the Living Arthropod image dataset were likely seen during pre-training. \autoref{tab:dev-secondary-common} and \autoref{tab:dev-secondary-rare} support this idea with high performance in more common taxa classification and poorer performance classifying more rare taxa. The RAG models did not see the same sharp decrease in accuracy and confidence as the Na\"ive LLM and Na\"ive VLMs in the rare taxa subset nor in the Rare Species subset \autoref{tab:rs-performance}, signalling the RAG model could be a viable method for rare and unknown species identification, particularly as more curated text-based knowledge sources become available. 

As citizen science initiatives develop and species trait characterization become more comprehensive, there is potential for our RAG pipeline to harness those improvements to better assist in identification and characterization of unknown species. Additionally, LLMs have recently been recruited to mine and curate data sources for biodiversity research \citep{dsouza_mining_2025} which would automate data curation and could soon boost the relevancy of RAG systems in Arthropoda taxonomy. 

Our RAG based approach to taxonomic classification represents a promising new paradigm that leverages explicit textual descriptions rather than relying solely on visual patterns. While our current implementation using Wikimedia and Wikispecies data showed modest improvements over the Na\"ive LLM, the approach has significant untapped potential. Even our efforts to improve the embedding space using semantic filtering of uninformative source chunks and contextualizing text did not yield marked improvements, suggesting that the current limiting factor is the breadth and depth of curated taxonomic text data, particularly at the genus and species levels. Alternative embedding structures such as GraphRAG \citep{han_retrieval-augmented_2025} could improve retrieval, as demonstrated in other biological domains \citep{li_grappi_2025}. Additionally, many models have demonstrated the classification benefits of multimodal over unimodal embedding spaces \citep{gong_clibd_2024, zhang_biomedclip_2023, stevens_bioclip_2023}, suggesting that RAG systems could similarly benefit from multimodal embeddings of images paired with text. 

Our approach first converts visual features into text through image captioning, creating an abstraction layer for taxonomic classification. While our prompting strategy reduced inappropriate inferences during captioning (\autoref{appendix-captioning prompt}), developing an organismal image-caption dataset with domain experts could enable fine-tuning of VLMs specifically for biological image captioning. Fine-tuning would help ensure captions contain only visually-evident features and could provide a tool to generate image-biocaption datasets. A fine-tuned captioner would, on its own, be valuable for automated trait descriptions of potentially novel taxa (i.e.~BINs), aiding in both indexing of operational taxonomic units and functional inference. These detailed descriptions would in turn contribute to expanding the contextual knowledge bases available for taxonomic classification. 

Recent work has shown VLMs can exhibit shape and texture biases \citep{gavrikov_are_2024}, potentially limiting their inference to species seen during pre-training. By reasoning over explicit visible traits in captions combined with retrieved contextual knowledge, our RAG approach can better identify novel species compared to direct image classification. This advantage is particularly evident through evaluation on the Rare Species subset of uncommon species and could be extended by evaluation on the full Rare Species dataset~\citep{stevens_bioclip_2023}. Additionally, incorporating human feedback during fine-tuning, which has proven successful for calibrating LLM confidence scores \citep{tian_just_2023}, could further improve the quality of our generated biodiversity knowledge.

\section{Conclusion}

We have demonstrated that pairing RAG models with VLMs offers a promising approach to taxonomic classification of unknown Arthropoda species, particularly as biodiversity knowledge bases continue to improve. While Na\"ive VLMs excelled in identifying familiar species, our RAG approach can leverage contextual information to classify rare and unknown taxa. The growing integration of citizen science data will further enhance species identification and inform conservation strategies. We envision future work expanding these models to more diverse datasets, especially those including underrepresented species. Our work establishes a novel bridge between modern AI tools and traditional text-based knowledge bases in biodiversity conservation, combining advanced RAG techniques, complex LLM-tooling, and LLM-based evaluation. By adapting and incorporating emerging AI techniques, we can significantly advance our understanding and conservation of biodiversity.

\section*{Acknowledgements and Disclosure of Funding}

Scott C.~Lowe provided valuable feedback on results visualizations and manuscript readability. Elizabeth Campolongo provided technical assistance for Rare Species experiments. This project was initiated within the Vector Institute RAG Bootcamp. We thank the organizers and the staff who supported its implementation.

This work was supported by the \href{http://abcresearchcenter.org/}{AI and Biodiversity Change (ABC) Global Center}, which is funded by the US National Science Foundation under \href{https://www.nsf.gov/awardsearch/showAward?AWD_ID=2330423&HistoricalAwards=false}{Award No. 2330423} and Natural Sciences and Engineering Research Council of Canada under \href{https://www.nserc-crsng.gc.ca/ase-oro/Details-Detailles_eng.asp?id=782440}{Award No. 585136}. We also acknowledge the support of the Government of Canada's New Frontiers in Research Fund under \href{}{Award No. NFRFT-2020-00073}. This research was supported, in part, by the Province of Ontario and the Government of Canada through the Canadian Institute for Advanced Research (CIFAR), and \href{https://vectorinstitute.ai/partnerships/current-partners/}{companies sponsoring} the Vector Institute.

GWT is supported by the Natural Sciences and Engineering Research Council of Canada (NSERC), the Canada Research Chairs program, and the Canada CIFAR AI Chairs program.

\section*{References}
\printbibliography[heading=none]

\appendix

\renewcommand{\thesection}{\Alph{section}}
\renewcommand{\thesubsection}{\thesection.\arabic{subsection}}

\section{Dataset Summary Tables}
\label{appendix-dataset tables}

\begin{table}[H]
\centering
\caption{Summary of taxonomic counts by rank in Living Arthropod dataset}
\begin{tabular}{|l|c|c|c|c|}
\hline
Class & Order Count & Family Count & Genus Count & Species Count \\
\hline
Arachnida & 3 & 16 & 27 & 32 \\
Chilopoda & 2 & 2 & 2 & 2 \\
Diplopoda & 3 & 3 & 3 & 4 \\
Insecta & 8 & 37 & 81 & 108 \\
Malacostraca & 2 & 7 & 7 & 7 \\
\hline
\end{tabular}
\end{table}

\begin{table}[H]
\centering
\caption{Summary of taxonomic counts by rank in Rare Species subset}
\begin{tabular}{|l|c|c|c|c|}
\hline
Class & Order Count & Family Count & Genus Count & Species Count \\
\hline
Arachnida & 1 & 2 & 2 & 3 \\
Branchiopoda & 1 & 1 & 1 & 1 \\
Insecta & 6 & 13 & 22 & 28 \\
Malacostraca & 1 & 1 & 1 & 1 \\
\hline
\end{tabular}
\end{table}

\section{Pipeline Prompts}
\label{appendix:prompts}

We present the prompts used in our pipeline below. Each prompt is structured to elicit specific outputs from the underlying models.

\subsection{Contextualizing Chunks Prompt}
\label{appendix-contextualize prompt}

\begin{promptbox}
\begin{prompttext}

You are an expert AI assistant to a taxonomist. You can determine if a user-provided chunk of textual context is useful for a taxonomist and if it is, contextualize it for retrieval in the context of a larger document.

You are analyzing text from documents about organisms. For each chunk:
1. First determine if the chunk (alone) contains actual descriptive content about organisms. A chunk with just a citation, header, references, or other non-descriptive text should be marked as not useful.
2. If the chunk contains descriptive content about organisms, please give a short succinct context to situate this chunk within the overall document for the purposes of improving search retrieval of the chunk. Also include taxonomic classification from the document.
3. If the chunk does NOT contain descriptive content, briefly state what type of content it contains instead (e.g. "citation", "references header", etc).

You do not need to give an overview starting statement such as "This document provides a detailed description of the phylum Chordata" but rather, get right into what is being said about Chordata in the chunk.

Format your response as a parsable JSON with:
- useful: boolean indicating if descriptive organism content is present
- contextual\_text: string containing either the descriptive contextualization or content type explanation

<document>
\{doc\_content\}
</document>

Here is the chunk we want to situate within the whole document

<chunk>
\{chunk\_content\}
</chunk>

\end{prompttext}
\end{promptbox}

\textit{Note: 256 thinking dot tokens are appended to the prompt above.}

\subsection{Module I: Image Captioning Prompt}
\label{appendix-captioning prompt}

\begin{promptbox}
\begin{prompttext}
You are an expert AI vision assistant to a taxonomist that describes animals in images.

Your task is to describe in extensive detail all the physical features (body and head shape, appendages, colour pattern, shape, texture, etc) of any organism(s) observable in the image.

Ensure each feature is elaborated upon wherever possible. Elaborate on the visual traits and morphology of subsections of the organism such as any appendages (such as limbs, wings etc) that are visible in the image — aim to describe them thoroughly.

Additionally, include a comprehensive description of the current state and/or life stage of the organism, with nuances in coloration, wear, or any distinctive features.

Please describe the environmental context surrounding the animal in detail. For example, if a butterfly is resting on flowers, you should also delve into the unique characteristics of the flowers, such as their shape, color, and arrangement.

Avoid using species common names (such as 'Monarch' for a butterfly).

Aim for a detailed analysis of at least 7 sentences with no upper limit if the detail demands it.

Do not include emotional descriptors (such as 'peaceful setting') or any other non-visible descriptors. Everything you mention should be evident in the image to an observer. Your response must rely solely on visual cues from the image, avoiding any inferences that are not evident.

Write a extremely detailed caption for the organism in the image and without commenting on the contrast or 'feeling' of the image.

An example of the type of caption you should produce is:
    Insecta with 4 visible jointed legs, partially translucent wings and compound eyes. There is a three-part body with a head, thorax and abdomen. An anterior lateral view of an adult fly with an abdomen that is mostly black and has a black tail-like taper. The wings have streaks of white as does the thorax and are black elsewhere. The prescutum and scutum are brown and in addition to the head, have small shiny hairs. The wings attach at the middle of the thorax, as do the legs. The legs have an initial black segment but are mostly coppery-brown and terminate into a triangular base. The wings are not as long as the length of the body and lay relatively flat at an angle away from the body with 2 segmented translucent halteres. The head is copper, orange and brown with white bordering. The head is visibly segmented from the thorax but the thorax and abdomen appear continuous and not visibly segmented. One brownish-orange eye with a white border is fully visible and the other eye is partially visible. There are two coppery kidney-shaped mouth parts protruding from the lower front of the head. A single shiny antennae is visible. The fly is standing on a green leaf that has pointed edges.

\end{prompttext}
\end{promptbox}

\textit{Note: 256 thinking dot tokens are appended to the prompt above.}

\subsection{Module II: RAG System Prompt}
\label{appendix-rag prompt}

\begin{promptbox}
\begin{prompttext}

You are an expert AI taxonomist. Your task is to use the organisms discussed in the caption of a new organism to generate a taxonomic classification for the new organism.

You will also be provided with some context that could or could not match the caption, if there is information in the context that matches the caption, you can use that info to inform your decision about the taxonomic classification, otherwise, if information does not match the details provided in the caption, disregard it.

You will be provided with context and a caption, provide in your response:
1. A Taxonomic classification
2. A description pairing physical traits common to both the new organism described in the caption and other organisms described in the context that indicate and support the choice made in the taxonomic classification.
3. A description of physical traits particular to this new organism described in the caption. These traits may set it apart from other organisms, may suggest it has unique features, and/or contain traits that may be candidates to investigate for a more specific taxonomic classification.
4. Commentary on your choice, including discussion of confidence, what new information about the new organism would help support the taxonomic classification and what new information would dispute the taxonomic classification.
5. A few paragraphs describing the features present (from the caption) and how they could relate to the biodiversity knowledge that is relevant to the taxa chosen.

Do not include a taxonomic classification for a certain rank unless you are confident from the caption (and/or context) about the classification.

<context>
\{context\}
</context>

<caption>
\{caption\}
</caption>

\end{prompttext}
\end{promptbox}

\textit{Note: 256 thinking dot tokens are appended to the prompt above.}

\end{document}